\documentclass{IOS-Book-Article}

\usepackage{mathptmx}
\usepackage[table]{xcolor}
\usepackage{soul}\setuldepth{article}
\usepackage{hyperref}
\usepackage{amsmath,amssymb}
\usepackage{graphicx}
 
\def\hb{\hbox to 10.7 cm{}}

\begin{document}

\pagestyle{headings}
\def\thepage{}

\begin{frontmatter}              % The preamble begins here.

%\pretitle{Pretitle}
\title{Sentence Embeddings and High-speed Similarity Search for Fast Computer Assisted Annotation of Legal Documents}

\markboth{}{December 2020\hb}
%\subtitle{Subtitle}

\author[A]{\fnms{Hannes} \snm{Westermann}
\thanks{Corresponding Author: Hannes Westermann, E-mail: hannes.westermann@umontreal.ca}},
\author[B]{\fnms{Jarom\'{i}r} \snm{\v{S}avelka}},
\author[C]{\fnms{Vern R.} \snm{Walker}},
\author[D]{\fnms{Kevin D.} \snm{Ashley}} and
\author[A]{\fnms{Karim} \snm{Benyekhlef}}

\runningauthor{H. Westermann et al.}
\address[A]{Cyberjustice Laboratory, Facult\'e de droit, Universit\'e de Montréal}
\address[B]{School of Computer Science, Carnegie Mellon University}
\address[C]{LLT Lab, Maurice A. Deane School of Law, Hofstra University}
\address[D]{School of Computing and Information, University of Pittsburgh}

\begin{abstract}
Human-performed annotation of sentences in legal documents is an important prerequisite to many machine learning based systems supporting legal tasks. Typically, the annotation is done sequentially, sentence by sentence, which is often time consuming and, hence, expensive. In this paper, we introduce a proof-of-concept system for annotating sentences ``laterally.'' The approach is based on the observation that sentences that are similar in meaning often have the same label in terms of a particular type system. We use this observation in allowing annotators to quickly view and annotate sentences that are semantically similar to a given sentence, across an entire corpus of documents. Here, we present the interface of the system and empirically evaluate the approach. The experiments show that lateral annotation has the potential to make the annotation process quicker and more consistent.
\end{abstract}

\end{frontmatter}
\markboth{December 2020\hb}{December 2020\hb}

\section{Introduction}
%Much AI \& Law research requires annotating legal texts. 
A lot of AI \& Law research is enabled by annotation of legal texts. The annotation can be performed on several levels of textual units, such as the entire document, the paragraph, or an individual sentence. In this work, we focus on annotations performed on the sentence level. AI \& Law research has employed a variety of annotation schemes on the sentence level, such as the annotation of:

%Further, the attribute of these textual units that is annotated can be of several different types, depending on the use-case. For example, a sentence might be labeled in terms of:
\begin{itemize}
    \item the rhetorical roles sentences play in a legal case (such as factual circumstances, a legal rule or an application of a legal rule to factual circumstances);
    \item the presence or absence of a certain factual circumstance the sentence describes (such as whether a security measure was present in a trade-secret case);
    \item the type and attributes of contractual clauses (such as the kind of liability addressed in a certain clause); and 
    \item the relevance of a sentence retrieved from a legal case to interpret a statutory term.
\end{itemize}

\noindent An annotated corpus of documents has many useful applications. For instance, a classification algorithm may be trained to infer labels for new sentences in a larger corpus of documents. This may lead, for example, to insights about the distribution of clauses in a large data set of contracts, to improved predictions about the outcome of a legal case from factors, or to ranking documents by relevance to a particular search query.

%Creating  annotated corpora  usually involves: (1) creating an annotation scheme to determine how sentences should be labeled, and (2) employing annotators to review legal textual documents and assigning labels to the documents. This paper  focuses on the second of these steps, how to assign labels on a sentence-basis to a set of documents. 

%\subsection{The traditional approach - sequential annotation}
Typically,  annotation of documents is performed by one or several annotators using a tool that allows them to review one document at a time, and to sequentially assign a label for each sentence as it occurs in that document. This approach has significant drawbacks. First, it is inefficient because annotating large corpora in this way takes a long time and is expensive. The label has to be determined from scratch for each sentence, causing significant cognitive overhead. 
Second, the annotations might be inconsistent across similar sentences. Since annotators often work through thousands of sentences, they may not remember how a certain sentence type was annotated the last time they saw it. If multiple annotators are involved, this problem may be exacerbated, as similar sentences are reviewed by different annotators.

%\subsection{A new approach - lateral annotation}
In this paper, we investigate an alternative approach which we call ``lateral annotation.'' Similarly to the traditional approach, annotators use a tool to view documents and label sentences. However, given any sentence there is an option to see sentences across the entire corpus (or from the rest of the document) that are semantically similar to the focused sentence. This feature uses sentence encoders based on deep learning models and libraries to quickly deliver semantically similar sentences using approximate nearest neighbour searches. The annotator then has the option of reviewing these similar sentences and assigning labels to one or more of them. Although the computer system assists the user by showing similar sentences, the choice of how to label a sentence ultimately rests with the annotator. It is therefore a hybrid approach, using machine learning to support human annotators with their task. 

Legal language is often formalized and uses recurring linguistic structures. This means that identical, or very similar, sentences often appear in many documents. For example, contract clauses specifying a certain type of liability might often use the same words, syntax and sentence structure. Lateral annotation makes use of this attribute of legal language by allowing the annotator to label all similar sentences at one time. This can increase the speed of annotation. Since all similar sentences can be labelled at once, the consistency of the annotations is also likely to increase. Consequently this approach can significantly ease the important task of labelling large data sets in the legal domain.

\section{Related Work}
Branting et al. \cite{branting2020,branting2019} proposed a semi-supervised approach to annotation of case decisions. The approach is based on several observations about the consistency of language across separate cases and within different sections of the same case. The researchers annotated a small set of decisions and calculated the mean of the semantic vectors \cite{bojanowski2017,joulin2016} of all the spans annotated by a given tag (the “tag centroid”). The annotations were then projected to semantically similar sentences in the entire corpus to enable explainable prediction. In our work, we describe a hybrid method where we show the semantically similar sentences to an annotator for rapid and reliable annotation.

A steady line of work in AI \& Law focuses on making the annotation effort more effective. Westermann et al. \cite{westermann2019} proposed and assessed a method for building strong, explainable classifiers in the form of Boolean search rules. Employing an intuitive interface, the user develops Boolean rules for matching instead of annotating the individual sentences. Here, we replace the Boolean matching rules with sentence semantic similarity. Instead of developing the rules, the user  confirms that the semantically similar sentences should be labeled as instances of the same types. \v{S}avelka and Ashley \cite{savelka2015} evaluated the effectiveness of an approach where a user labels the documents by confirming (or correcting) the prediction of a ML algorithm (interactive approach). The application of active learning has been explored in the context of classification of statutory provisions \cite{waltl2017} and eDiscovery \cite{cormack2016,cormack2015}. Hogan et al. \cite{hogan2009} proposed and evaluated a human-aided computer cognition framework for eDiscovery. Tools to retrieve and rank text fragment by similarity for coding have further been implemented in qualitative data analysis tools, such as QDA Miner\footnote{\url{provalisresearch.com/products/qualitative-data-analysis-software}} and Nvivio.\footnote{www.qsrinternational.com/nvivo-qualitative-data-analysis-software/home}

\section{Proposed Framework}
We investigate a system that enables an annotator to perform lateral annotations on a corpus of documents. We use sentence embeddings to capture the meaning of sentences, and then use approximate nearest neighbour search to find sentences that are semantically similar to a source sentence. This enables us to provide the annotators with viable sentence candidates for annotation in sub-second time.

\subsection{Sentence Embeddings} \label{embeddings}
In order to search for similar sentences based on an original sentence, we need a way to store sentences in a vector format that makes comparison easy. There are several ways of representing sentences in this way. 

A bag of words representation (e.g., TF-IDF) is a  simple but effective way to encode the meaning of a sentence. It has, however, at least two  notable disadvantages:  an enormously large feature space and the inability to account for the relatedness of meaning in different words. This means that sentences with the same meaning may be deemed to be completely different if they use largely non-overlapping vocabulary (e.g., synonyms). This is problematic for applications where sentence similarity is a key component (as in this work). In our experiments, we include the representation as a strong baseline  due to its simplicity and wide usage.

More recently, pre-trained word embeddings and language models have gained popularity in creating word embeddings. These representations are motivated by the so-called  distributional hypothesis: words that are used and occur in the same contexts tend to have similar meanings. \cite{harris1954} The idea that ``a word is characterized by the company it keeps'' was popularized by Firth. \cite{firth1957} The gist of the method is that words with similar meanings are projected onto similar vectors, by analyzing massive corpora of text to learn the distributions. There are several ways of combining these word vectors to produce sentence vectors, of which we have chosen three:

\begin{enumerate}
\item Cer et al. \cite{cer2018} use the transformer architecture \cite{vaswani2017} and Deep Averaging Network \cite{iyyer2015} trained on the SNLI dataset. We work with the implementation released by the authors as the Google Universal Sentence Encoder (\textbf{GUSE}).\footnote{\url{https://tfhub.dev/google/universal-sentence-encoder/4}} 
\item Reimers et al. \cite{reimers2019} build on top of BERT \cite{devlin2018} and RoBERTa \cite{liu2019}, which have been shown to be remarkably effective on a number of NLP tasks. Specifically, they use siamese and triplet network structures to derive semantically meaningful sentence embeddings. The authors released the models as Sentence Transformers (\textbf{ST}). We use this implementation in our work.\footnote{\url{github.com/UKPLab/sentence-transformers}}
\item Conneau et al. \cite{conneau2017} demonstrate the effectiveness of models trained on a natural language inference task (Stanford NLI dataset \cite{bowman2015}). They propose a BiLSTM network with max pooling trained with fastText word embeddings \cite{bojanowski2017,joulin2016} as the best universal sentence encoding method. We adapt the implementation released by the authors which is commonly  referred to as \textbf{InferSent}.\footnote{\url{github.com/facebookresearch/InferSent}}
\end{enumerate}

\subsection{Efficient Similarity Search over High-dimensional Vectors}

Document search traditionally relies on a combination of relational databases built on structured data (metadata search) and inverted indexes (full-text search). These cannot  deal efficiently  with the vectors that represent documents' meaning. The brute-force approaches to indexing (i.e., store the results for all documents) or querying (i.e., compare the query to each data point) do not scale well beyond fairly small data sets.

In order to achieve  semantic similarity search with the desired properties, one has to tackle the problem of preprocessing a set of $n$ data points $P=\{p_1, p_2 \ldots, p_n\}$ in some metric space $\mathbb{X}$ (e.g., the d-dimensional Euclidean space $\mathbb{R}^d$) so as to efficiently answer queries by finding the point in $P$ closest to a query point $q\in\mathbb{X}$. Solutions to this well-studied problem in other domains can be readily applied in our context. \cite{indyk1998} For example, images and videos have become a massive source of data for indexing and search. And since it is often not practical to manually annotate the data to enable the use of relational databases, a search in some sort of a vector space remains the only option.

Johnson et al. \cite{johnson2019} proposed a system that allows efficient indexing and search over collections containing around 1 billion documents. This shows that the current state-of-the-art is capable of supporting virtually any practical scenario in a legal annotation domain. For example, \v{S}avelka et al. \cite{savelka2019} segmented the whole corpus of US case-law into 0.5 billion sentences. While the technique in \cite{johnson2019} would allow for efficient semantic similarity search over such a collection, it is most likely several magnitudes larger than any realistic legal annotation task. We utilize the Annoy similarity search library released by Spotify\footnote{\url{github.com/spotify/annoy}}  for its ease of use and minimal system requirements. Annoy is an efficient implementation of the Approximate Nearest Neighbors algorithm proposed in \cite{indyk1998}.

\subsection{Lateral Annotation}
\begin{figure}
    \includegraphics[width=1\textwidth]{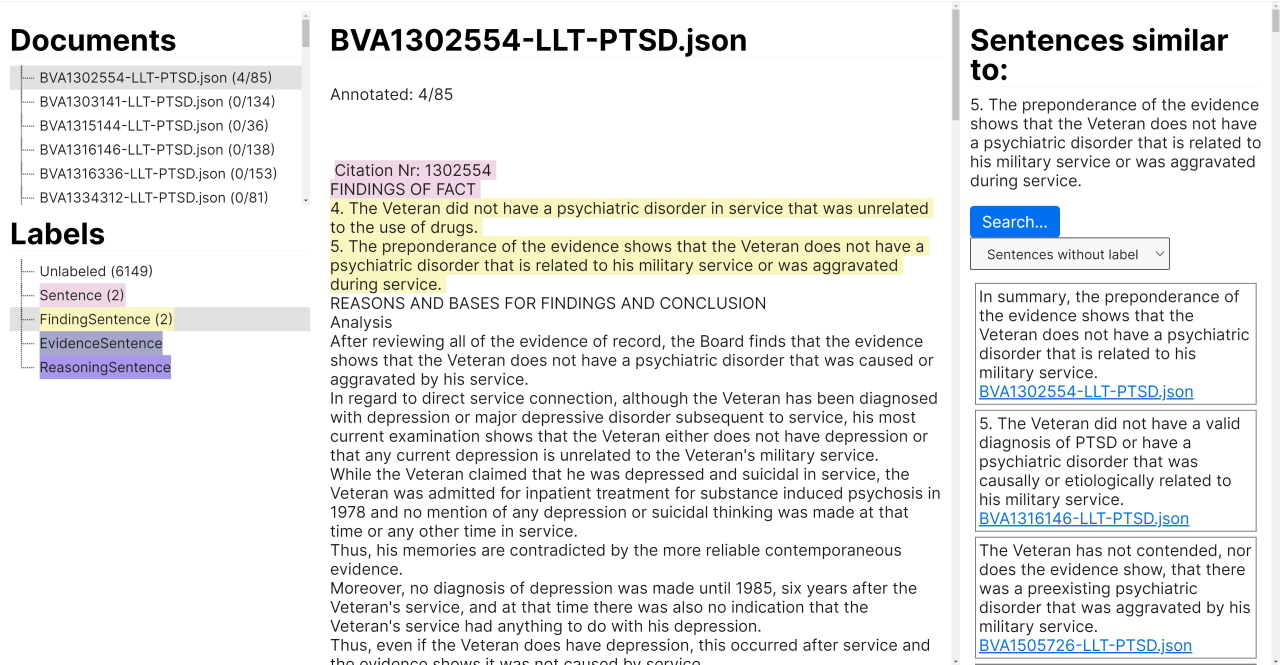}
    \caption{A screenshot of the prototype Computer-Assisted Efficient Semantic Annotation \& Ranking (CAESAR) Interface}
    \label{fig:caesar}
\end{figure}

Semantic sentence embeddings and efficient vector similarity search are combined to enable lateral annotation. We have developed a prototype interface called CAESAR, Computer-Assisted Enhanced Semantic Annotation \& Ranking, to demonstrate this capability. The sentence embedding frameworks are used to create semantic embeddings of all sentences in a corpus. These are then used to create an index for fast similarity search.

The capability is provided to the user through an annotation interface (see Figure \ref{fig:caesar}). The interface  allows the annotator to define a schema of labels in a hierarchical structure (i.e. a type system), and to tag individual sentences with these labels. For each sentence, it is possible to retrieve semantically similar sentences, using the methods described above. These are shown in the sidebar to the right, sorted by similarity in descending order. This panel allows the annotator to perform lateral annotations by quickly annotating the sentences shown, or to see the context of the shown sentences before annotation.

We envision the following procedure for labeling sentences using CAESAR. An annotator starts with the first document, and labels the first sentence. Then, he asks to be shown similar sentences in the sidebar. He then labels sentences in the sidebar until sentences are no longer similar enough to quickly allow the annotator to determine that they are of the same class. The annotator then returns to the full text of the case and labels the next sentence. As the annotator moves to the next documents, many of the sentences may already be labelled, and can be skipped.

This method of improving annotation efficiency is completely unsupervised. It can be implemented before having started any kind of annotation, by relying on the sophisticated neural models trained on huge datasets of general texts (e.g., news corpora, Wikipedia). Despite this, the method seems to perform very well on legal annotation tasks, as demonstrated in Section \ref{sec:eval}.

\section{Evaluation}
\label{sec:eval}

\subsection{Datasets}
In order to evaluate the lateral annotation method, we use three existing data sets:

\begin{enumerate}
\item Walker et al. \cite{walker2019} analyzed 50 fact-finding decisions issued by the U.S. Board of Veterans' Appeals (\textbf{BVA}) from 2013 through 2017, all arbitrarily selected cases dealing with claims by veterans for service-related post-traumatic stress disorder (PTSD). For each of the 50 BVA decisions in the PTSD dataset, the researchers extracted all sentences addressing the factual issues related to the claim for PTSD, or for a closely-related psychiatric disorder. These were tagged with the rhetorical roles  the sentences play in the decision \cite{walker2017}. We conducted our experiments on this set of 6,153 sentences.\footnote{Dataset available at \url{github.com/LLTLab/VetClaims-JSON}}
\item \v{S}avelka et al. \cite{savelka2019} studied methods for retrieving useful sentences from court opinions that elaborate on the meaning of a vague statutory term. To support their experiments they queried the database of sentences from case law that mentioned three terms from different provisions of the U.S. Code. They manually classified the sentences in terms of four categories with respect to their usefulness for the interpretation of the corresponding statutory term. In \cite{savelka2019} the goal was to rank the sentences with respect to their usefulness; here, we classify them into the four value categories (\textbf{StatInt}).\footnote{Dataset available at \url{github.com/jsavelka/statutory_interpretation}}
\item Bhattacharya et al. \cite{bhattacharya2019} analyzed 50 opinions of the Supreme Court of India. The cases were sampled from 5 different domains in proportion to their frequencies (criminal, land and property, constitutional, labor and industrial, and intellectual property). From each of the 50 decisions the sentence boundaries were detected using an off-the-shelf general tool.\footnote{\url{spacy.io}} Then the 9,380 sentences were manually classified into one of the seven categories according to the rhetorical roles they play in a decision. Our experiments were conducted on this set of sentences (\textbf{IndSC}).\footnote{Dataset available at \url{github.com/Law-AI/semantic-segmentation}}
\end{enumerate}

\subsection{Experiments} \label{experiments}
In order to evaluate the effectiveness of lateral annotation and compare different embedding methods, we use our system to retrieve the closest sentences to a query sentence, and investigate how many of them have the same label as the source sentence. Assuming that lateral annotation is more efficient than sequential annotation, the more retrieved sentences that have the same label, the more efficiently the annotator will be able to annotate the data set.

We report several metrics. First, we investigate the length of chains of annotations by traversing the retrieved sentences, from the most similar to the least, until we arrive at a label that does not match the label of the source sentence. We calculate the longest encountered chain (Max) and the average chain length (Avg) for each data set and embedding method. Second, we determine how many of the top 20 closest sentences have the same label as the source sentence (P@20) - a measure of precision.

Third, we visualize the high-dimensional GUSE embeddings of all sentences in a dataset, reduced to 2 dimensions using a Principal Component Analysis. The colors in the resulting visualization correspond to the gold standard labels for the individual sentences. The most important feature of an embedding space for our purpose is that each sentence should be surrounded by multiple sentences with the same label, e.g. the same color.

\subsection{Results}

%\subsubsection{General Analysis}
Table \ref{tab:stats} presents the Max, Avg, and P@20 statistics for each data set and for each embedding method. Overall, the sentence embeddings seem to capture enough linguistic information to achieve improvement in all three metrics, without any training on the domain-specific data sets. The neural models perform much better than the random baseline, and perform better or equal to the TF-IDF baseline. 

%\subsubsection{Dataset analysis}
Looking at the individual data sets, it seems like the \textbf{Board of Veterans' Appeals data set} benefits significantly from lateral annotation. On average, 70\% of the closest 20 sentences to each sentence have the same label, meaning an annotator can likely annotate these sentences laterally. This could offer a significant speed-up in the annotation of such a data set. 

\begin{figure}
    \includegraphics[width=.3\textwidth]{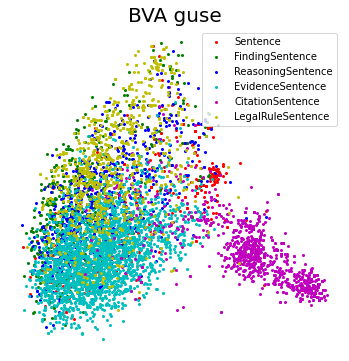}
    \includegraphics[width=.3\textwidth]{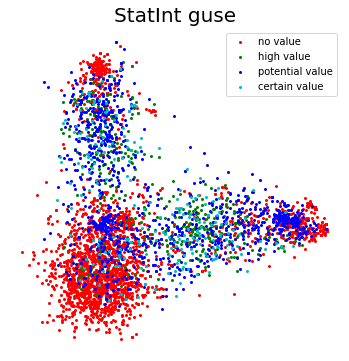}
    \includegraphics[width=.3\textwidth]{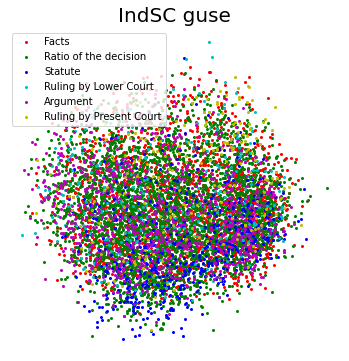}
    \caption{Visualizations of the sentences across the data sets, embedded using the GUSE and reduced to two dimensions using a Principal Component Analysis. The colors correspond to different labels.}
    \label{fig:cluster_plots}
\end{figure}

Looking at the individual labels in Table \ref{tab:per_label}, the ``Citation'' label seems by far the easiest to annotate laterally. 94\% of the top 20 closest sentences to a citation sentence are also citation sentences. This is likely due to the special tokens (such as parentheses, year numbers and special words such as ``See'')  in these sentences. ``Rule'' also performs very well, which might be due to the same rule being cited in multiple cases. The embeddings capture these distinctions well, which can be seen in Figure \ref{fig:cluster_plots}, where sentences of the same type seem clearly concentrated in certain areas.

In the \textbf{Statutory Interpretation data set} the technique appears most suitable for the sentences labeled as ``No value.'' This makes sense since these are mostly sentences that fully or partially quote or paraphrase a statutory provision. Hence, these sentences are often very similar to each other. The middle graph in Figure \ref{fig:cluster_plots} confirms this observation. Three compact red clusters clearly correspond to the ``No value'' sentences associated with the three terms of interest. The sentences with the other three labels are somewhat more challenging. Yet, even for the more challenging categories, a significant amount of sentences could still be annotated laterally, as seen in Table \ref{tab:per_label}.

The \textbf{Indian Supreme Court data set} is where lateral annotation gives the least advantage, with our models retrieving under 40\% of matching sentences in the top 20 positions. On average, each sentence seems to be next to only 2.1-2.4 sentences of the same class in the embedding space. This can also be seen in the PCA visualization in figure \ref{fig:cluster_plots}. Unlike the other data sets, the sentence embeddings do not seem to result in clearly separate classes. The comparative difficulty of separating this data set might be explained by the fact that the sentences are selected from five different domains, and assigned seven labels---more than the other two data sets. The ``Ratio'' and ``Facts'' sections seem slightly easier to annotate in a lateral fashion, which might be due to a consistent structure or content of these sentences. It is surprising that the ``Ratio'' class has a high precision, while the ``Ruling of lower court'' has low precision, but this matches the findings of the authors in \cite{bhattacharya2019} for classification difficulty.

\begin{table}[]
    \centering
    \setlength{\tabcolsep}{6pt}
    \begin{tabular}{lrrr|rrr|rrr}
              & \multicolumn{9}{c}{\cellcolor{black!8}Statistics} \\
              & \multicolumn{3}{c|}{BVA} & \multicolumn{3}{c|}{StatInt} & \multicolumn{3}{c}{IndSC} \\
              & Max & Avg  & P@20 & Max  & Avg  & P@20 & Max  & Avg  & P@20 \\
              \hline
    Random    &   9 &  1.35& .24   &   18 &     2& .45   &   10 & 1.36 & .24   \\ 
	TF-IDF      & 195 & 13.73& .59   &  197 & 24.40& .70   &   27 & 2.16 & .36  \\
	GUSE      & 696 & 55.92& \textbf{.70} &  257 & 25.80 &\textbf{.73}&\textbf{45}&\textbf{2.43}&.37\\ 
	ST        & 710 & 48.62& .68   &\textbf{277}&\textbf{30.16}& .69   &   30 & 2.22 & .35   
	\\ 
	InferSent &\textbf{863}&\textbf{83.92}& \textbf{.70}   &  204 & 22.5 & .66   &\textbf{45}& 2.41 & \textbf{.38}   \\ 
    \end{tabular}
    \caption{Statistics for different sentence embedding methods, including evaluation of chains of lateral annotation (Max, Avg) and how many of the 20 closest sentences on average have the same label (P@20).}
    \label{tab:stats}
\end{table}

\begin{table}[]
    \centering
    \setlength{\tabcolsep}{9pt}
    \begin{tabular}{lr|lr|lr}
              \multicolumn{2}{c|}{\cellcolor{black!8}BVA} & \multicolumn{2}{c|}{\cellcolor{black!8}SID} & \multicolumn{2}{c}{\cellcolor{black!8}ISC} \\
              
              Label & P@20 & Label & P@20 & Label & P@20\\
              \hline
                Sentence &  0.60 & No Value & 0.89 & Facts & 0.40 \\
                Finding &  0.49 & Potential & 0.58 & Ruling (lower) & 0.08 \\
                Evidence &  0.76 & Certain & 0.15 & Argument & 0.18 \\
                Rule &  0.73 & High & 0.33 & Ratio & 0.46 \\
                Citation &  0.94 & &  & Statute & 0.25 \\
                Reasoning &  0.27 &  &  & Precedent & 0.32 \\
                &   &  &  & Ruling (present) & 0.32 \\
	 
    \end{tabular}
    \caption{The ratio of matching labels among the top 20 most similar sentences, per label}
    \label{tab:per_label}
\end{table}

\section{Discussion}
We have introduced and evaluated a lateral annotation framework. We experimented with four types of sentence embeddings and compared them against the random baseline. All of the embeddings showed significant improvements in selecting sentences that are of the same class compared to the random baseline. The neural models show strong performance across the three data sets. In general, they perform similarly, although the Google Universal Sentence Encoder and InferSent seems to have a slight edge. In the BVA data set, the neural models clearly outperform the TF-IDF baseline, while the performance is more balanced in the StatInt and IndSC data sets. Even in the most challenging data set, almost 40\% of the 20 closest sentences had the same label as a source sentences. For the other data sets, this number was 70\%. This indicates a significant potential benefit for using lateral annotation.

Different areas might benefit from the use of lateral annotations. The assumption behind the method is that sentences that have similar semantic embeddings are likely to belong to the same class that an annotator is aiming to label. This should work better for labeling schemas and data sets where the semantic properties of a sentence are linked to its label, and where 
the homogeneity of sentences with the same label is high. This can be seen in the per-class analysis of precision, showing that citation sentences and recitation of previous rules and cases are more suitable for lateral annotation. Sentences with less inherent structure and similarity, such as reasoning sentences, seem to perform slightly worse. The Indian Supreme Court data set, which draws from five different domains and uses seven classes, performs worse in a lateral annotation context, which could indicate that more diverse data sets are more difficult to annotate laterally. 

Further, the method benefits from data sets where the set of sentences with a particular label is made up of several clusters of semantically similar sentences that the annotator can  efficiently scope. For each sentence that is part of such a cluster, the annotator can efficiently label a large number of sentences. Outlier sentences, which do not belong to any larger cluster of sentences with the same label, are less likely to benefit from the method, as they do not assist the annotator in finding other sentences of the same label.

We hypothesize that the legal domain is well-suited for the lateral annotation method. Legal practitioners often use stereotypical language to describe certain facts, including a shared vocabulary and sentence structure. This shared language is more likely to be suited for annotation supported by semantic similarity search, and could significantly speed-up  annotating large data sets with per-sentence labels.

\section{Future Work}
There are multiple ways of building upon this research. First, it is important to investigate how lateral annotation performs in additional real-world scenarios, and compare it to traditional methods of annotation. Second, finding ways to expand our framework by extending vectors with relevant properties or by combining vectors could increase the system's performance. Third, augmenting the method to integrate active learning  approaches (where a machine learning model suggests which sample to label next to the annotator) could help to discover more relevant sentences. Furthermore, the approach of annotating sentences laterally could be used at an earlier stage, to support the exploration of data sets and the creation of type systems.

\section{Conclusions}
In this paper, we have explored a method for the efficient annotation of sentences, by leveraging sophisticated sentence embedding models and approximate nearest neighbour searches. Using these technologies, we designed a method and an interface that allow annotators to label similar sentences in one go across documents, rather than having to episodically label similar sentences as they come up in new documents. We investigated some properties of different possible embeddings and demonstrated the benefits of using the method on three legal data sets.

\end{document}